\documentclass{ieeeaccess}
\usepackage{cite}
\usepackage{amsmath,amssymb,amsfonts}
\usepackage{algorithmic}
\usepackage{graphicx}
\usepackage{textcomp}

\usepackage{times}

\usepackage{bbold}
\usepackage{bm}
\usepackage{multirow}
\usepackage{nccmath}
\usepackage[driverfallback=dvipdfm]{hyperref}
\hypersetup{
setpagesize=false,
 bookmarksnumbered=true,%
 bookmarksopen=true,%
 colorlinks=true,%
}

\newcommand*{\eg}{\textit{e.g.}}
\newcommand*{\ie}{\textit{i.e.}}
\newcommand*{\etal}{\textit{et al.}}

\def\BibTeX{{\rm B\kern-.05em{\sc i\kern-.025em b}\kern-.08em
    T\kern-.1667em\lower.7ex\hbox{E}\kern-.125emX}}

\begin{document}
\history{Received June 11, 2021, accepted June 28, 2021.}
\doi{10.1109/ACCESS.2021.3094052}
\title{Foreground-Aware Stylization and Consensus Pseudo-Labeling for Domain Adaptation of First-Person Hand Segmentation}

\author{\uppercase{Takehiko Ohkawa}\authorrefmark{1,2}, \uppercase{Takuma Yagi}\authorrefmark{1}, \uppercase{Atsushi Hashimoto}\authorrefmark{2}, \IEEEmembership{Member, IEEE}, \uppercase{Yoshitaka Ushiku}\authorrefmark{2}, \IEEEmembership{Member, IEEE}, and \uppercase{Yoichi Sato}\authorrefmark{1}, \IEEEmembership{Senior Member, IEEE}}

\address[1]{Institute of Industrial Science, The University of Tokyo, Tokyo 153-8505, Japan  (e-mail: \{ohkawa-t, tyagi, ysato\}@iis.u-tokyo.ac.jp)}
\address[2]{OMRON SINIC X Corporation, Tokyo 113-0033, Japan  (e-mail: \{atsushi.hashimoto, yoshitaka.ushiku\}@sinicx.com)}
\tfootnote{
This work was supported in part by JST ACT-X Grant Number JPMJAX2007, JST AIP Acceleration Research Grant Number JPMJCR20U1, and JST KAKENHI Grant Number 17H06100, Japan.
This work was also supported in part by a hardware donation from Yu Darvish.}


\corresp{Corresponding author: Takehiko Ohkawa (e-mail: ohkawa-t@iis.u-tokyo.ac.jp).}

\begin{abstract}
Hand segmentation is a crucial task in first-person vision. 
Since first-person images exhibit strong bias in appearance among different environments, adapting a pre-trained segmentation model to a new domain is required in hand segmentation.
Here, we focus on appearance gaps for hand regions and backgrounds separately.
We propose (i) foreground-aware image stylization and (ii) consensus pseudo-labeling for domain adaptation of hand segmentation.
We stylize source images independently for the foreground and background using target images as style.
To resolve the domain shift that the stylization has not addressed, we apply careful pseudo-labeling by taking a consensus between the models trained on the source and stylized source images.
We validated our method on domain adaptation of hand segmentation from real and simulation images. Our method achieved state-of-the-art performance in both settings.
We also demonstrated promising results in challenging multi-target domain adaptation and domain generalization settings. Code is available at \href{https://github.com/ut-vision/FgSty-CPL}{https://github.com/ut-vision/FgSty-CPL}.
\end{abstract}


\titlepgskip=-15pt

\DeclareRobustCommand\red{}
\DeclareRobustCommand\blue{}
\DeclareRobustCommand\magenta{}
\def\Hline{\noalign{\hrule height 0.4mm}}
\newcommand{\hyphen}{\mathchar`-}
\def\br{\ \\}
\def\bla{bla bla bla bla bla bla bla bla bla bla bla bla bla bla bla bla bla bla bla bla bla bla bla bla bla bla bla bla bla bla bla bla bla bla bla bla bla bla bla bla bla bla bla bla bla bla}
\def\tbf{\textbf}
\def\tit{\textit}
\def\bs{\boldsymbol}
\def\cvpr{Proceedings of the IEEE Conference on Computer Vision and Pattern Recognition (CVPR)}
\def\cvprw{Proceedings of the IEEE Conference on Computer Vision and Pattern Recognition Workshops (CVPRW)}
\def\ijcv{Proceedings of the International Journal of Computer Vision (IJCV)}
\def\iccv{Proceedings of the IEEE International Conference on Computer Vision (ICCV)}
\def\eccv{Proceedings of the European Conference on Computer Vision (ECCV)}
\def\icml{Proceedings of the International Conference on Machine Learning (ICML)}
\def\nips{Proceedings of the Advances in Neural Information Processing Systems (NeurIPS)}
\def\iclr{Proceedings of the International Conference on Learning Representations (ICLR)}

\def\TY#1{\textcolor{red}{(TY: #1)}}

\def\fig#1#2{
\begin{figure}[!t]
\centering
\includegraphics[width=0.9\hsize]{fig/#1}
\caption{#2}
\label{fig:#1}
\end{figure}
}

\def\figw#1#2{
\begin{figure*}[!t]
\centering
\includegraphics[width=0.9\hsize]{fig/#1}
\caption{#2}
\label{fig:#1}
\end{figure*}
}

\def\tableDA{
\begin{table}[!t]
\caption{Real-to-real and sim-to-real adaptation of hand segmentation. EGTEA, Ego2Hands, and ObMan-Ego are used for the source dataset. We report average mean IoU (\%) across all the target domains~\cite{gtea,edsh,utg,yhg}.}
\vspace{-2pt}
\centering{
\scalebox{1}[1]{
\begin{tabular}{l|ccc}\Hline
Method                     & EGTEA & Ego2Hands & ObMan-Ego \\\hline
Source only                & $65.47$ & $36.44$ & $25.94$     \\\hline
PL                         & $57.35$ & $47.80$ & $44.32$     \\
BDL~\cite{bdl}             & $67.18$ & $28.22$ & $36.66$     \\
UMA~\cite{uma}             & $73.41$ & $37.25$ & $7.43$     \\\hline
Proposed (FgSty)               & $74.35$ & $60.83$ & $61.05$     \\
Proposed (CPL)                 & $68.70$ & $46.10$ & $45.14$     \\
Proposed (FgSty + CPL)         & $\bm{74.95}$ & $\bm{68.74}$ & $\bm{65.79}$     \\\hline\hline
Target only                & $85.14$ & $85.14$ & $85.14$     \\
\Hline
\end{tabular}
}}
\label{tbl:da}
\end{table}
}

\def\tableMTDA{
\begin{table}[!t]
\caption{Single-target adaptation vs. multi-target adaptation. EGTEA, Ego2Hands, and ObMan-Ego are used for the source dataset. We report average mean IoU (\%) across all the target domains~\cite{gtea,edsh,utg,yhg}.}
\centering
\scalebox{0.82}[1]{
\begin{tabular}{lc|ccc}\Hline
Method                     & Target & EGTEA & Ego2Hands & ObMan-Ego \\\hline
Proposed (FgSty + CPL)         & Single & $74.95$ & $68.74$ & $65.79$     \\\hline
Proposed (FgSty + CPL)         & Multiple  & $74.97$ & $70.07$ & $63.60$ \\
Proposed (FgSty + CPL + Adv)   & Multiple  & $\bm{76.20}$ & $\bm{70.83}$ & $\bm{70.00}$\\
\Hline
\end{tabular}
}
\label{tbl:mtda}
\end{table}
}

\def\tableAdv{
\begin{table}[!t]
\caption{Integration with adversarial adaptation.  EGTEA, Ego2Hands, and ObMan-Ego are used for the source dataset. We report average mean IoU (\%) across all the target domains~\cite{gtea,edsh,utg,yhg}.}
\centering
\scalebox{0.95}[1]{
\begin{tabular}{l|ccc}\Hline
Method                     & EGTEA & Ego2Hands & ObMan-Ego \\\hline
Source only                & $65.47$ & $36.44$ & $25.94$     \\
Proposed (FgSty)               & $74.35$ & $60.83$ & $61.05$     \\
Proposed (CPL)                 & $68.70$ & $46.10$ & $45.14$     \\
Proposed (FgSty + CPL)         & $74.95$ & $68.74$ & $65.79$     \\\hline
Source only + Adv          & $66.44$ & $49.00$ & $53.22$     \\
Proposed (FgSty + Adv)         & $74.43$ & $70.71$ & $71.38$     \\
Proposed (CPL + Adv)           & $62.74$ & $52.63$ & $27.32$     \\
Proposed (FgSty + CPL + Adv)   & $\bm{75.23}$ & $\bm{70.76}$ & $\bm{72.01}$ \\
\Hline
\end{tabular}
}
\label{tbl:adv}
\end{table}
}

\def\tableDG{
\begin{table}[!t]
\caption{Domain generalization of hand segmentation. ObMan-Ego is used for the source dataset. Mean IoU (\%) is used for evaluation. G, E, U, and Y denote GTEA, EDSH, UTG, and YHG, respectively. "Proposed" shown in the Table denotes our full model; Proposed (FgSty + CPL).}
\centering{
\scalebox{1}[1]{
\begin{tabular}{c|c|cc||c}\Hline
Test & {Source only} & Auxiliaries & Proposed  & Target only \\\hline
GTEA & $6.05$ & E, U, Y & $62.72$ & $91.97$ \\
EDSH-2 & $33.67$ & G, U, Y & $70.70$ & $84.23$ \\
EDSH-K & $29.40$ & G, U, Y & $70.98$ & $76.85$ \\
UTG & $44.97$ & G, E, Y & $80.17$ & $90.81$ \\
YHG & $15.64$ & G, E, U & $14.69$ & $81.84$ \\\hline\hline
Avg. & $25.94$ & -- & $59.85$ & $85.14$ \\\Hline
\end{tabular}
}}
\label{tbl:dg}
\end{table}
}

\def\tableUnalign{
\begin{table}[!t]
\caption{Mask-aligned stylization vs. unaligned stylization. EGTEA is used for the source dataset. Mean IoU (\%) is used for evaluation. G, E2, EK, U, and Y denote GTEA, EDSH-2, EDSH-K, UTG, and YHG, respectively.}
\centering{
\scalebox{0.87}[1]{
\begin{tabular}{l|ccccc||c}\Hline
Condition & G & E2 & EK & U & Y & Avg.\\\hline
Source only & $\bm{89.45}$ & $73.53$ & $74.87$ & $59.62$ & $29.87$ & $65.47$\\
Unaligned & $87.98$ & $\bm{77.27}$ & $75.08$ & $61.85$ & $32.50$ & $66.94$\\
Proposed (FgSty) & $89.09$ & $71.86$ & $\bm{76.35}$ & $\bm{80.62}$ & $\bm{53.82}$ & $\bm{74.35}$\\\Hline
\end{tabular}
}}
\label{tbl:align}
\end{table}
}

\def\tableNorm{
\begin{table}[!t]
\caption{Comparison to image normalization methods. We implemented simple baselines of image normalization: gray-scaling, histogram equalization (HE), feature distribution matching (FDM)~\cite{keepsimple}, and color histogram matching (HM)~\cite{keepsimple}. EGTEA is used for the source domain. We report average mean IoU (\%) across all the target domains~\cite{gtea,edsh,utg,yhg}.}
\centering{
\scalebox{1}[1]{
\begin{tabular}{cccc|c}\Hline
Gray-scaling & HE & FDM~\cite{keepsimple} & HM~\cite{keepsimple} & Proposed (FgSty)\\\hline
$58.36$ & $66.21$ & $68.61$ & $66.63$ & $74.35$\\
\Hline
\end{tabular}
}}
\label{tbl:norm}
\end{table}
}

\def\tableSpeed{
\begin{table}[!t]
\caption{\red{Computational time of training with pseudo-labeling. We calculated the update speed (ms/iteration) with NVIDIA Tesla V100. We allocated the two networks in the consensus pseudo-labeling to two GPUs while the comparison methods used a single GPU. "Naive" is the training of a model on a single dataset without any adaptation.}}
\centering{
\scalebox{1}[1]{
\begin{tabular}{c|cc|c}\Hline
Naive & PL & UMA~\cite{uma} & Proposed (CPL)\\\hline
$394$ & $653$ & $841$ & $614$\\
\Hline
\end{tabular}
}}
\label{tbl:speed}
\end{table}
}

\def\tableSSize{
\begin{table}[!t]
\caption{\red{Ablation study of the source dataset size. EGTEA is used for the source dataset. We report average mean IoU (\%) across all the target domains~\cite{gtea,edsh,utg,yhg}. "Proposed" shown in the Table denotes our full model; Proposed (FgSty + CPL).}}
\vspace{-2pt}
\centering{
\scalebox{1}[1]{
\begin{tabular}{c|cc}\Hline
Source size & {Source only} & Proposed \\\hline
13K (full)  & $65.47$ & $74.95$\\
5k          & $65.52$ & $74.43$\\
3k          & $62.44$ & $71.83$\\
1k          & $58.77$ & $72.23$\\
500         & $45.17$ & $71.65$\\
\Hline
\end{tabular}
}}
\label{tbl:ssize}
\end{table}
}

\maketitle

\section{Introduction}\label{sec:intro}
Mobile cameras have become popular thanks to advances in photography, and a massive number of videos are recorded nowadays.
In particular, first-person vision~\cite{kanade2012first}, which captures interactions from the user's point of view by utilizing body-worn cameras, is gaining interest.
Analyzing first-person videos offers the opportunity for assisting in people's daily life and is useful in various applications, such as assistive technology~\cite{lee2020hand}, augmented reality~\cite{handsegar}, visual lifelogging~\cite{bolanos2016toward}, and human-robot interaction~\cite{wang2020see}.

When analyzing first-person videos, hands play a fundamental role in understanding the wearer's action and intention.
In particular, segmenting the hand region is crucial for several downstream tasks~\cite{fpvsurvey} such as hand pose estimation, 3D hand shape reconstruction, and hand-object interaction recognition.
A segmentation model must correctly segment the hand regions of diverse users and environments.

When deploying a pre-trained model to videos of a new user, domain shift between training and testing data is a significant problem in first-person vision.
The surrounding environments, appearance, viewpoints, and activities vary among different users. The data distributions among each user are also biased.
For example, the appearance of the hand regions and backgrounds is diverse among different environments.
Annotating new labels for every user is unrealistic since the annotation cost is high in segmentation tasks.
We therefore need to adapt the segmentation model to a new user's environment. Thus far, few attempts have been made to address this problem~\cite{uma}.

\fig{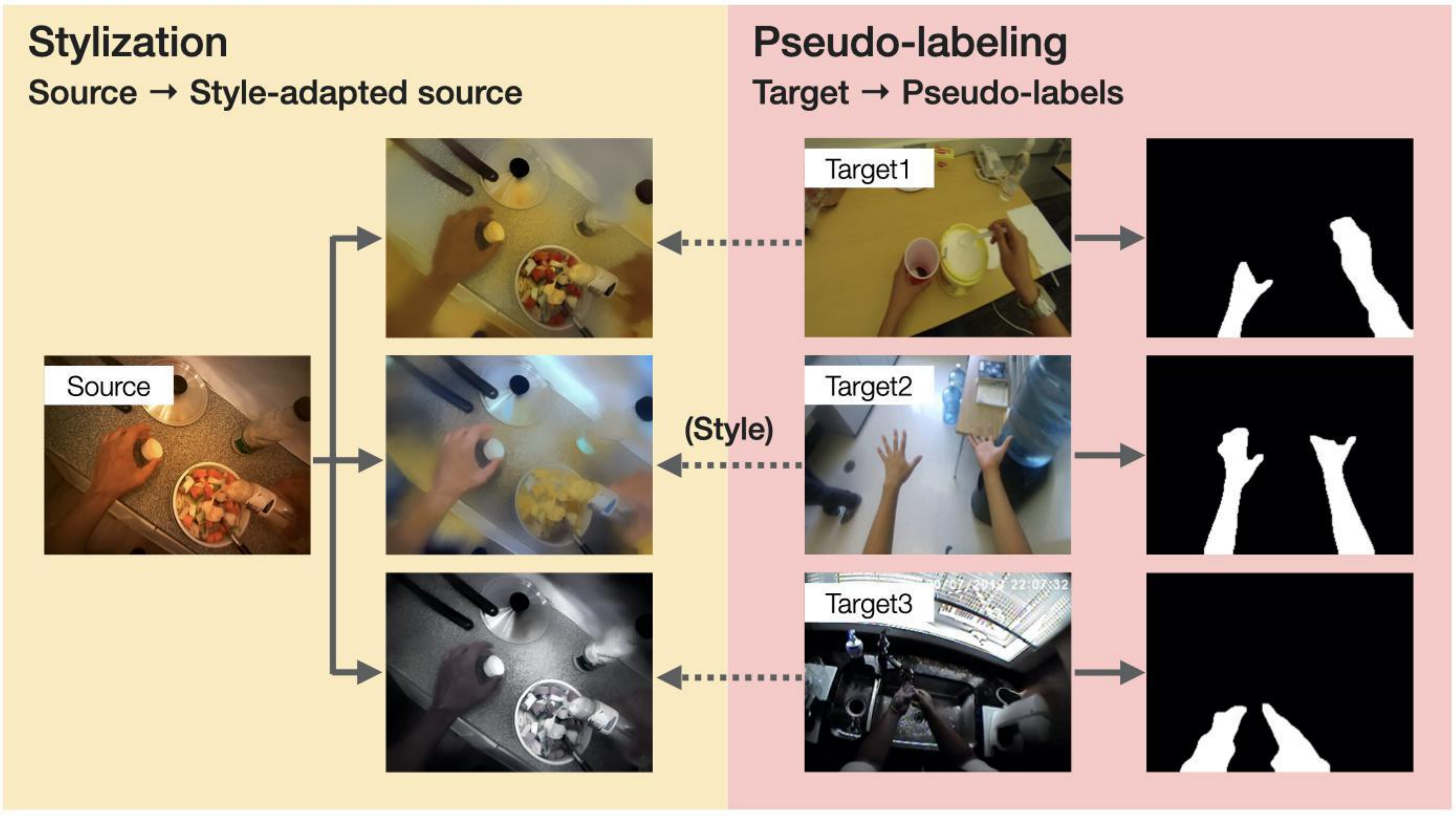}{Our two-stage domain adaptation approach. First, the stylization synthesizes a source image with the style of various target appearances and produces style-adapted source images. Second, the segmentation model learned on the style-adapted images assigns pseudo-labels to target data, which are utilized to further adapt the model to the target domain.}

To handle the domain shift in first-person vision, we propose a two-stage semi-supervised domain adaptation method for hand segmentation with (i) foreground-aware stylization and (ii) consensus pseudo-labeling (Fig.~\ref{fig:teaser3}). 
This work focuses on different levels of appearance gaps for the foreground (hands) and background in hand segmentation (Fig.~\ref{fig:fgbg}).
The first stylization approach stylizes style images as target styles separately for the foreground and background.
The second pseudo-labeling performs further feature-level alignment between the stylized source and target domain.

Our foreground-aware stylization generates source images adapted to the target domain's style by using style transfer~\cite{photowct}.
Assuming a few labeled target images for the target domain, we separately stylize the foreground and background regions using the target images as style images.
The style-adapted source images will be used to train the segmentation model. 

However, naively learning on the style-adapted source images is biased to the source label distribution, thus the model may not generalize to the target domain when its label distribution is different from the source domain. A simple solution is to fine-tune the model in the target domain.

To directly learn target discriminative representations, our pseudo-labeling assigns reliable pseudo-labels on target data, and the segmentation model is trained on the target images and pseudo-labels. 
When generating pseudo-labels, it is essential to assign them only when the prediction is confident.
While previous methods~\cite{bdl,cbst,styda,uma} have used their own model's confidence, since their thresholding to generate pseudo-labels is sensitive to a problem setting, their methods would be problematic in the first-person vision setting where the domain shift is caused by complex factors (\eg, differences in surrounding environments, appearance, viewpoints, and user's activities).

We therefore propose a discreet consensus pseudo-labeling technique that generates pseudo-labels based on the predictions of different segmentation abilities.
Specifically, we generate pseudo-labels by taking the intersection of the prediction of two networks trained on the source and style-adapted datasets, respectively.
This method reduces false assignments and carefully updates the segmentation network while preventing failures.

\fig{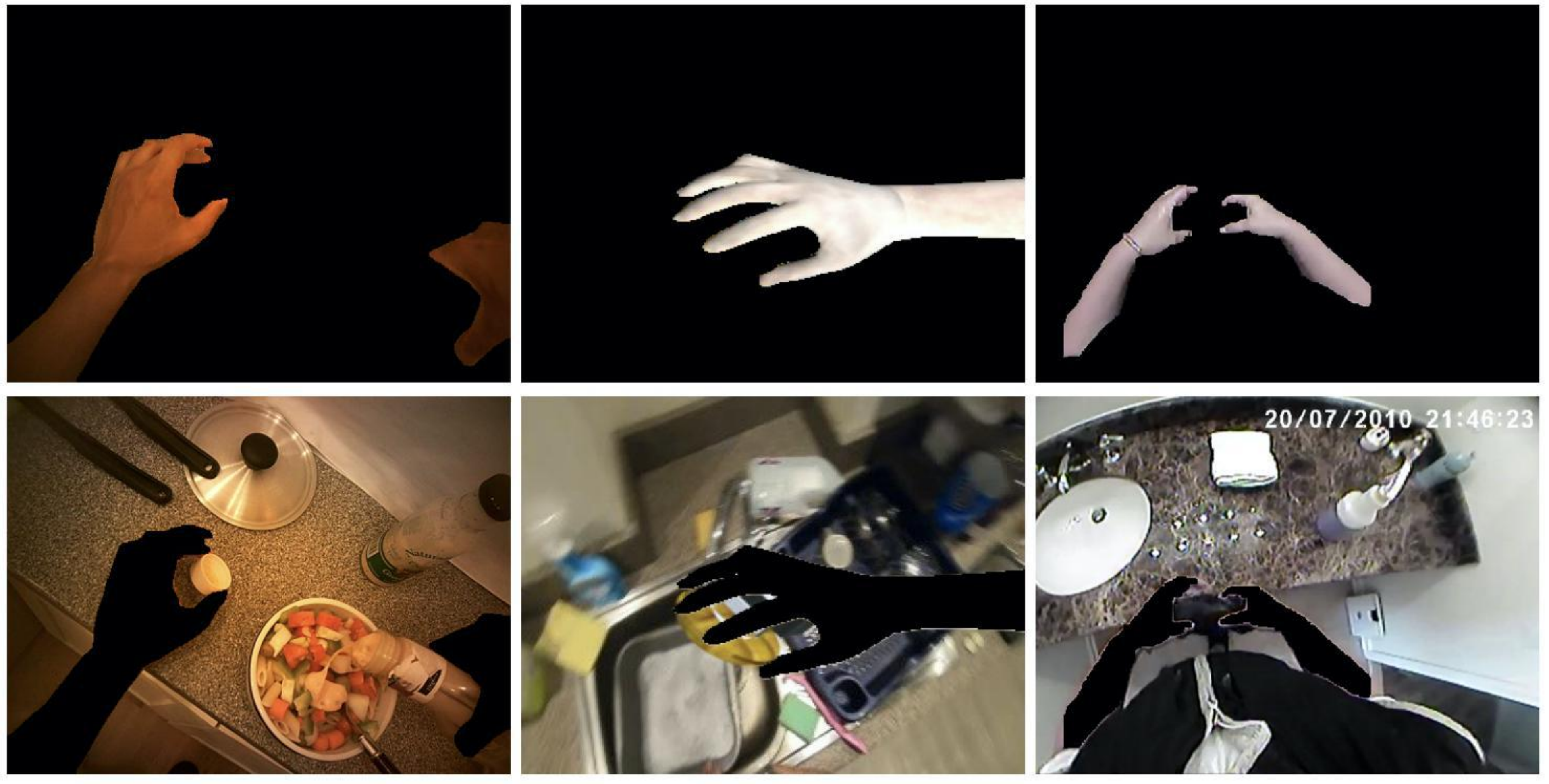}{Appearance-level domain gap. Foreground appearance (top) relies on skin color, light refection on hands, and presence or absence of synthetic texture. Background appearance (bottom) depends on camera configurations, objects in environment, and room light.}

In our experiments, we address domain adaptation tasks in hand segmentation where a real dataset~\cite{egtea} and two synthetic datasets~\cite{ego2hands} are used for the source dataset, and four first-person vision datasets~\cite{gtea,edsh,utg,yhg} are prepared for the target dataset. Our method improved mean IoU by $9.48\%$, $32.30\%$, and $39.85\%$ on average compared to unadapted performances when using EGTEA~\cite{egtea}, Ego2Hands~\cite{ego2hands}, and the newly created ObMan-Ego as the source dataset, respectively. Our method performed effectively in the sim-to-real setting with a large domain shift, where the state-of-the-art method~\cite{uma} cannot perform well.

We additionally explore further challenging adaptation and generalization scenarios.
When the target domain changes by the user's movement, adapting to multiple target domains is desired. 
When we cannot access the target data due to privacy issues~\cite{privacyeye,privacylpsensor,privacylr}, generalizing to an unseen domain is also necessary.
We also experimented on a multi-target domain adaptation setting~\cite{mtdainfo,blendmtda}, which aims to simultaneously adapt to multiple target domains, and a multi-auxiliary domain generalization setting~\cite{stydg}, which aims to generalize to an unseen test domain utilizing knowledge from auxiliary domains.

Our contributions are summarized as follows.
\begin{itemize}
    \setlength{\parskip}{0pt}
    \setlength{\itemsep}{5pt}
    \item To reduce the appearance gap of first-person images, we propose foreground-aware stylization to translate target appearance to source data independently for the foreground and background.
    \item We propose a consensus pseudo-labeling method based on the agreement of two networks trained on the source images and style-adapted source images produced by the stylization, respectively.
    \item We achieved state-of-the-art performance on domain adaptation of hand segmentation from real and simulation images compared to the latest domain adaptation methods.
\end{itemize}

\section{Related Work}~\label{sec:relate}
\tbf{Hand segmentation in first-person videos.} Hand segmentation is a task to segment the pixels of hands in an image. Recently, CNN-based hand segmentation has achieved impressive results. Bambach \etal~\cite{egohands} proposed a two-stage method that first detects hand bounding boxes with CNN and then segments the hand region by using GrabCut~\cite{grabcut}. Urooj and Borji~\cite{handwild} adopted an end-to-end approach using RefineNet~\cite{refinenet}, which has a multi-path refinement module that directly exploits the fine-grained features of shallow convolutional layers. However, the network is still sensitive to the domain shift in first-person videos~\cite{edsh,handwild}, such as changes in the surrounding environments, appearance, viewpoints, and the camera wearer's activities.

\tbf{Domain adaptation.} Most domain adaptation techniques are based on a single-target assumption, where the learning is performed from one domain to another. 
Domain adaptation methods are divided into three categories: (1) minimization of the distance between the source and target feature distributions~\cite{dann,adda,rtn,cada,mcdda,strongweak}, (2) generative approaches~\cite{cycada,learnfromsyn,dcan,bdl,fgbgda}, and (3) self-training~\cite{asymtri,cbst,styda,uma,advent,intrada}, which primarily exploits pseudo-labels for adaptation.
Specifically, in first-person vision, Cai~\etal~\cite{uma} proposed uncertainty-guided model adaptation for hand segmentation that iteratively assigns a pseudo-label from a confident prediction. 
Similar to our work, \cite{coda} and \cite{asymtri} estimate the uncertainty by the disagreement of two networks trained on a single source dataset to select reliable pseudo-labels. However, their works confirm the approach for classification only and extending it to segmentation is non-trivial. Lin~\etal~\cite{fgbgda} focused on the domain gaps separately for the foreground and background and proposed mask-aware segmentation loss and discriminator. 

In this work, we validate the agreement-based pseudo-labeling in segmentation.
Unlike~\cite{coda} and~\cite{asymtri}, both networks are trained by different training procedures and source datasets to determine pseudo-labels based on different viewpoints. This scheme prevents the generation of unreliable pseudo-labels where the two networks with similar feature representations make the same false prediction on the same region of an unlabeled instance and accept it as a final pseudo-label used for training.
Unlike~\cite{fgbgda}, we introduce foreground and background separation in the stylization. 

\tbf{Style transfer.}
Photo-realistic style transfer~\cite{deepphoto,photowct} is a type of style transfer~\cite{stytrans,demysty,adain} that modifies the visual style of a photo without distorting image edges. It produces photo-realistic images in a broad range of scenarios, including transitions of the time of day, weather, and season. 

Style transfer can be applied to domain adaptation by stylizing a source image as a target image's style~\cite{demysty}. \cite{fgbgda,domainsty} adopted the photo-realistic style transfer for semantic segmentation. 
Kim and Byun~\cite{styda} proposed texture-diversified style transfer to learn texture-invariant representation to diminish the gap between a synthetic and a target domain. Yue~\etal~\cite{stydg} used style transfer to learn domain-invariant representation for domain randomization.

In this work, we propose utilizing the photo-realistic style transfer~\cite{photowct} to alleviate appearance gaps separately for the foreground and background of first-person images.

\section{Proposed Method}\label{sec:method}
\fig{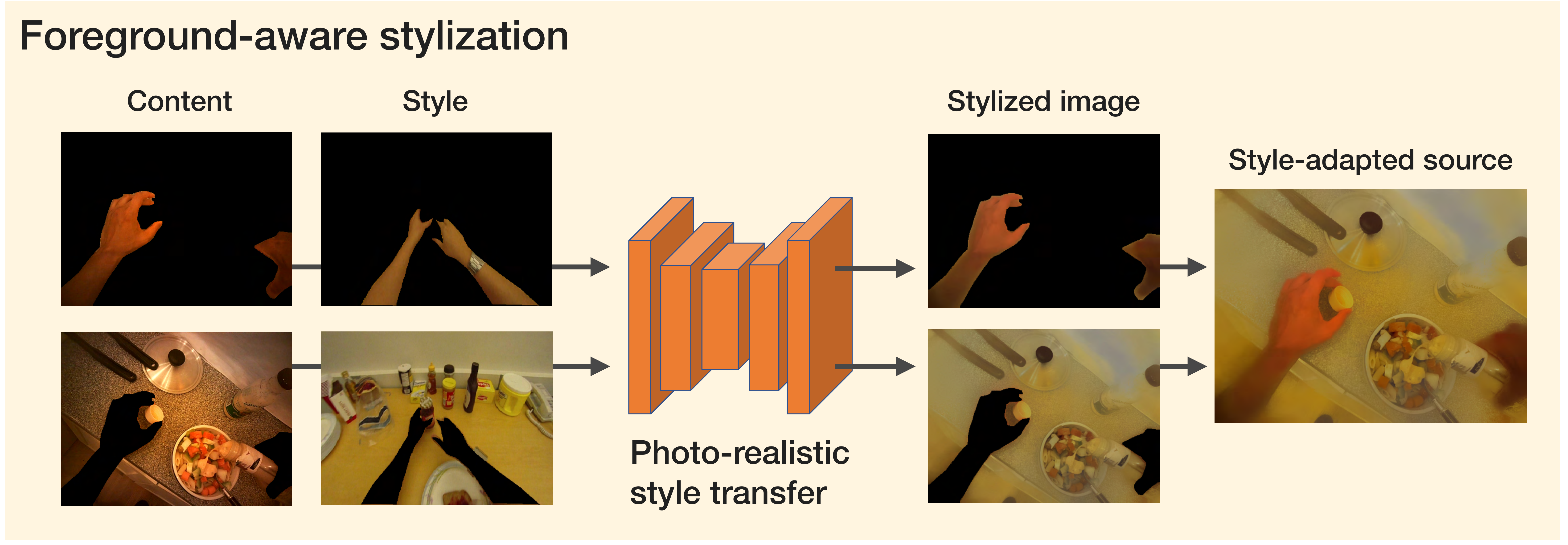}{Overview of our foreground-aware stylization. The style transfer stylizes the source data as the target styles while separating the foreground and background.}
\fig{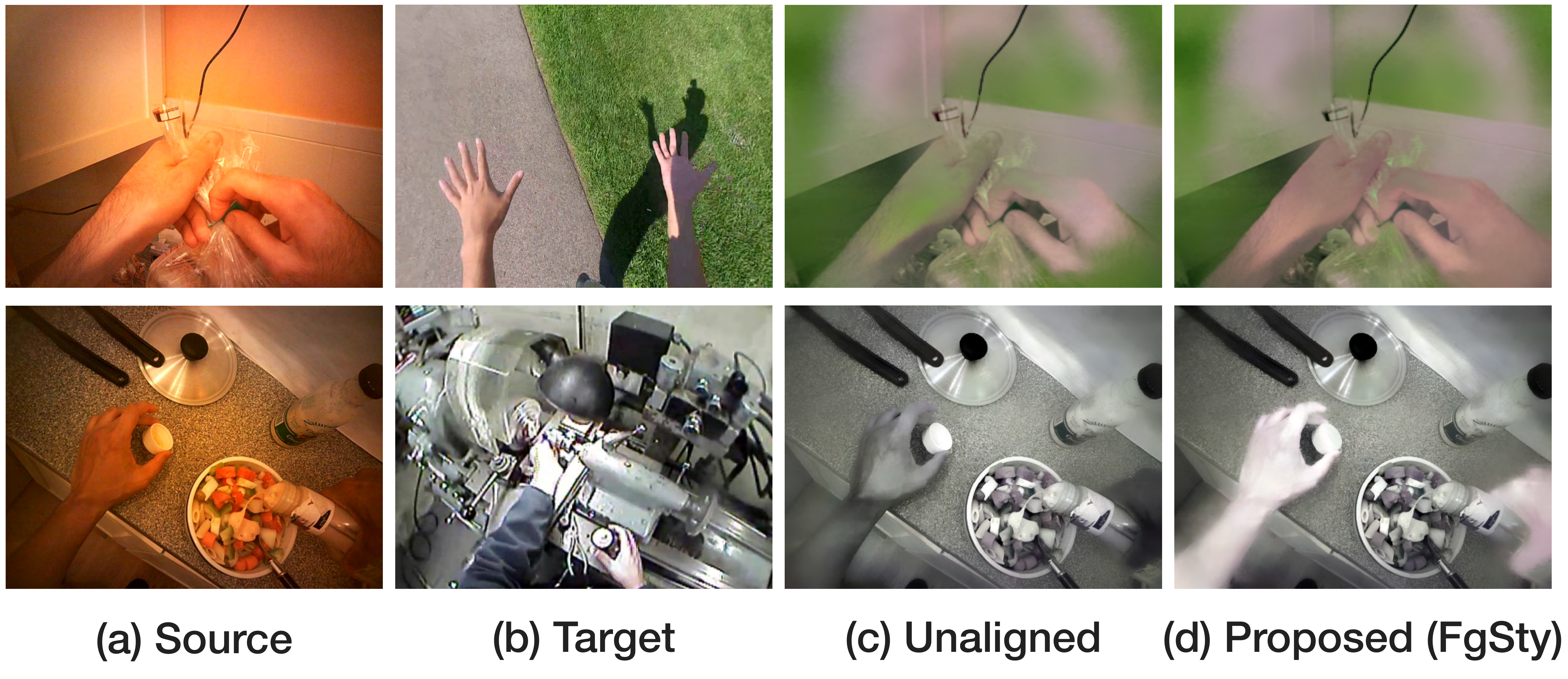}{Effect of mask alignment in stylization. While (c) the hands in the unaligned stylization were affected by the target backgrounds, (d) the aligned stylization faithfully transferred the target hand appearance.}

In this section, we present our proposed two-stage semi-supervised domain adaptation method.
We provide the details of our foreground-aware stylization and consensus pseudo-labeling.

Given a source dataset $\mathcal{S}$ and a target dataset $\mathcal{T}$, the goal of domain adaptation is to train a segmentation model using source images and labels $\{\bs{x}^{(s)}, \bs{y}^{(s)}\} \sim \mathcal{S}$ and target images $\{\bs{x}^{(t)}\} \sim \mathcal{T}$. The model is finally evaluated on test data of the target dataset.

\figw{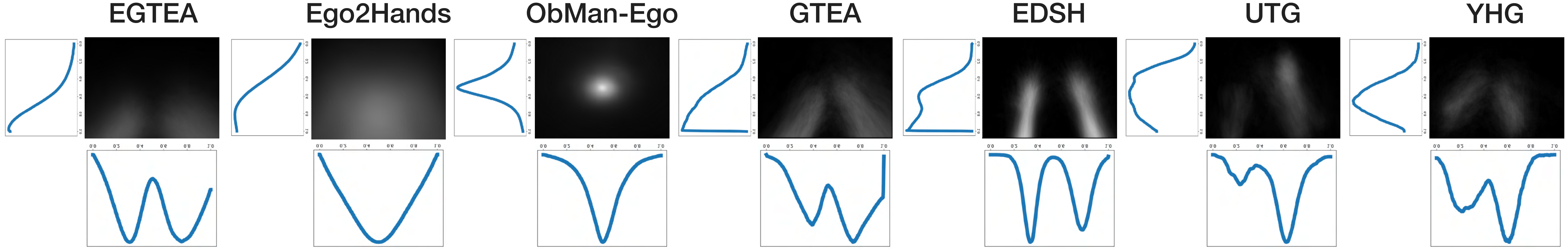}{Visualization of label distributions in first-person hand segmentation. Each figure represents an averaged hand mask over labels in the training dataset (center), and marginal distributions of the x-axis (bottom) and the y-axis (left) in the labels.}
\begin{figure*}[!t]
\centering
\includegraphics[width=0.98\hsize]{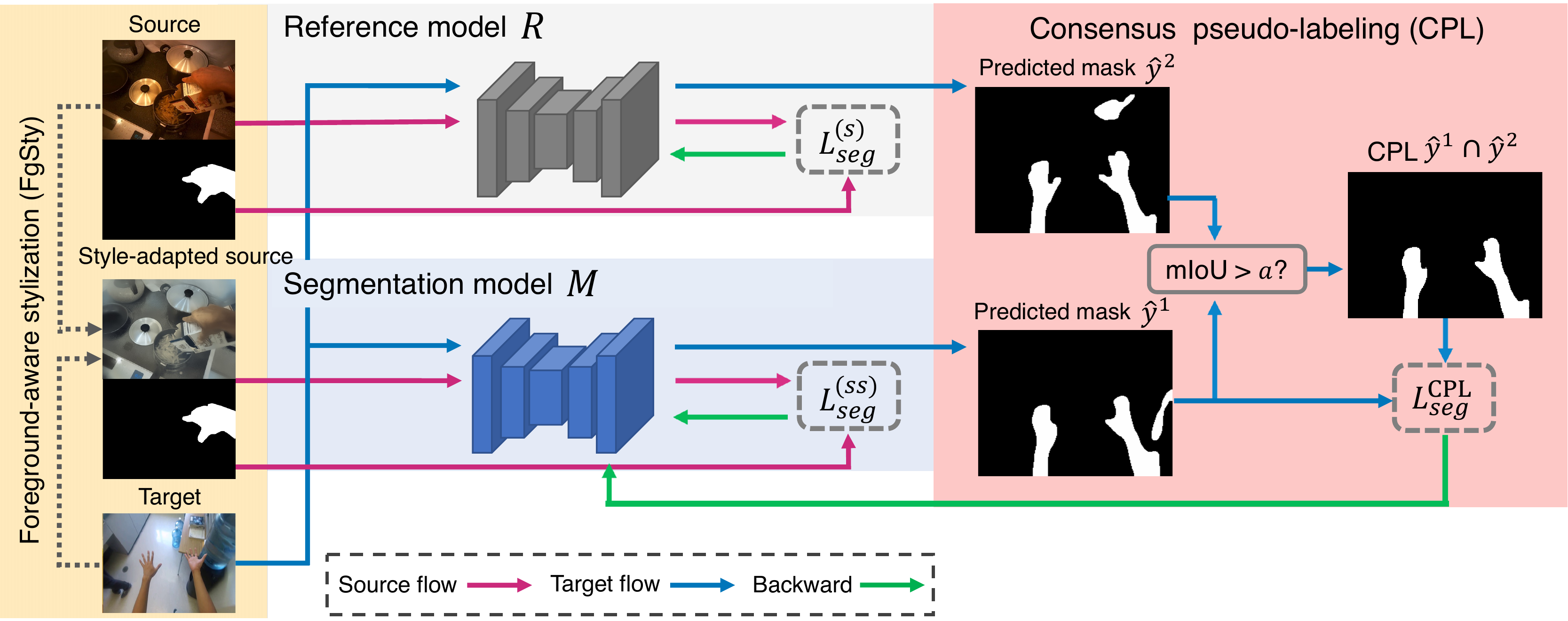}
\caption{Method overview. The reference model $R$ is learned on the source dataset $\mathcal{S}$ with the segmentation loss $L_{seg}^{(s)}$.
The segmentation model $M$ is learned to adapt from the style-adapted source dataset $\mathcal{SS}$ to the target dataset $\mathcal{T}$. In addition to the segmentation loss, $M$ is updated by the loss of the consensus pseudo-labeling $L_{seg}^{\mathrm{CPL}}$. The segmentation model is used for final prediction in testing.}
\label{fig:model}
\end{figure*}

\subsection{Foreground-Aware Stylization}\label{sec:fgsty}
Our first step is to synthesize style-adapted source images that have a similar appearance to the target dataset~\cite{fgsty}.
We propose using photo-realistic style transfer~\cite{photowct} with foreground-background separation, which produces a geometrically consistent image of a content image with the style of a style image.

Following the spatial control~\cite{stytrans} that allows users to control the content–style correspondences, we assume that a few target images are available with hand masks and separately stylize the foreground and background regions of a source image as the target images' style by using their masks. Fig.~\ref{fig:fgsty} illustrates the image synthesis process.

Fig.~\ref{fig:align} shows the results of the stylization with and without the separated stylization. 
In the unaligned case without the spatial control based on hand masks, the lighting of the hand region was affected by the background region of the target image. \red{While our aligned stylization may generate semantically inconsistent backgrounds, such as confusing the color of the road and the grass in the top result of Fig.~\ref{fig:align}~(d), the stylization successfully transferred the color of the target hand regions to the style-adapted source image.
Thus, the stylized images with the incorrectly aligned backgrounds do not affect the task of segmenting hands. 
In addition, since our second-stage pseudo-labeling can directly learn the target appearance, such partially misaligned backgrounds are acceptable in the first-stage stylization and expected to be corrected in the later training in the target domain.
}

Using this stylization with a few target images, we synthesize a style-adapted source dataset $\{\bs{x}^{(ss)}, \bs{y}^{(s)}\} \sim \mathcal{SS}$ with the source label $\bs{y}^{(s)}$.
We use this style-adapted dataset to train the segmentation network.

\subsection{Consensus Pseudo-Labeling}
Our second step is to train the segmentation model $M$ by using the style-adapted source dataset $\mathcal{SS}$. Nevertheless, it is not straightforward in hand segmentation. Naive training on the style-adapted images and source labels is biased to their label distribution. The learned model may not generalize to the target domain due to a spatial mismatch of label distributions between the source and target domain (\ie, target shift~\cite{targetcondshift,targetcondtrans,targetvae}). As illustrated in Fig.~\ref{fig:labelDist}, the problem of the label (hand) distribution shift is caused by differences in the user's activities and viewpoints. For instance, camera wearers work with both hands on EGTEA~\cite{egtea}, GTEA~\cite{gtea}, and EDSH~\cite{edsh} or heavily use their right hand on UTG~\cite{utg}. The angle of view is also very narrow~\cite{egtea} or tilted on YHG~\cite{yhg}.

A simple way to address the target shift is to fine-tune the model in the target domain. Since target data are unlabeled in domain adaptation, we adopt a pseudo-labeling approach, which assigns pseudo-labels $\hat{\bs{y}}^{(t)} = \mathbb{1}_{[M(\bs{x}^{(t)})>t]}$ to unlabeled target samples $\bs{x}^{(t)}$ by thresholding the prediction of the model $M$ by a fixed value $t$, and the pseudo-labels are used to update the model. However, since the choice of the threshold $t$ is sensitive to a problem setting, previous pseudo-labeling methods~\cite{bdl,cbst,styda,uma} may not work well in tasks with different source-to-target pairs.

To generate reliable pseudo-labels carefully, we propose consensus pseudo-labeling that exploits segmentation abilities attained in the stylized dataset and unstylized dataset (Fig.~\ref{fig:model}).
We introduce a reference model $R$ trained on the source dataset $\mathcal{S}$ and generate pseudo-labels by the agreement over predictions of the model $M$ trained on $\mathcal{SS}$ and the reference model $R$ trained on $\mathcal{S}$.
Given predicted masks $\hat{\bs{y}}^{1} = \mathbb{1}_{[M(\bs{x}^{(t)})>0.5]}$ and $\hat{\bs{y}}^{2} = \mathbb{1}_{[ R(\bs{x}^{(t)})>0.5]}$ for target data $\bs{x}^{(t)}$, we assign a pseudo-label if $\mathrm{mIoU}(\hat{\bs{y}}^{1}, \hat{\bs{y}}^{2}) > \alpha$, where $\mathrm{mIoU}$ returns the value of mean IoU between two inputs.
We determine the final pseudo-label as the intersection of the two masks $\hat{\bs{y}}^{(t)} = \hat{\bs{y}}^{1} \cap \hat{\bs{y}}^{2}$.
Using this label, we take a binary cross-entropy loss $L_{seg}^{\mathrm{CPL}}$ to update the model $M$. 

Since both models are trained by the stylized and unstylized source datasets and by different learning procedures, we expect the segmentation model $M$ and the reference model $R$ to learn different feature representations and predict an instance based on different viewpoints, as studied in~\cite{sin}.
As a result, the false prediction of $M$ can be suppressed by the prediction of $R$ and the quality of our final pseudo-label $\hat{\bs{y}}^{(t)}$ would be higher than $\hat{\bs{y}}^{1}$, the pseudo-label generated by the segmentation model $M$ only.

\subsection{Overall Objective}
Finally, we combine the proposed foreground-aware stylization and consensus pseudo-labeling. The full objective of the segmentation model $M$ consists of the segmentation loss $L_{seg}^{(ss)}$ on the style-adapted source dataset $\mathcal{SS}$ and the loss of consensus pseudo-labeling $L_{seg}^{\mathrm{CPL}}$ on the target data, which is formulated as
\begin{align}
    \min _{M} L_{seg}^{(ss)}(M) + L_{seg}^{\mathrm{CPL}}(M, R).
    \label{eq:full}
\end{align}
The reference model $R$ is trained with the segmentation loss $L_{seg}^{(s)}$ on the source dataset $\mathcal{S}$, which is written as
\begin{align}
    \min _{R} L_{seg}^{(s)}(R).
\end{align}
\section{Experiments}
In this section, we begin by introducing our experimental setup and then provide the main results, ablation studies, and additional experiments.
Specifically, we conducted experiments on adapting from real images (real-to-real adaptation) and synthetic images (sim-to-real adaptation). In the ablation studies, we explored the effect of our foreground-aware stylization and consensus pseudo-labeling.
We also report three additional results for our method (i) with adversarial adaptation, (ii) in a multi-target domain adaptation setting, which simultaneously adapts to multiple target domains by one-time training, and (iii) in a multi-auxiliary domain generalization setting, where we have several reference datasets but no access to the test domain.
In the experiments, we report mean Intersection over Union (mIoU) for evaluation.

\subsection{Experimental Setup}
\tbf{Datasets.}
Following~\cite{uma}, we used multiple first-person video datasets with various types of illumination.
We selected EGTEA~\cite{egtea} as a source dataset in the real-to-real setting.
In the sim-to-real setting, we used the large-scale synthetic datasets Ego2Hands~\cite{ego2hands} and ObMan-Ego as the source dataset. Using the rendering pipeline proposed in~\cite{obman}, we rendered the ObMan-Ego consisting of synthetic hands~\cite{graspit} and objects~\cite{shapenet} with the backgrounds of two large-scale egocentric videos, EPIC-KITCHENS-100~\cite{epic100} and Something-Something~\cite{something}. 
We prepared GTEA~\cite{gtea}, EDSH~\cite{edsh}, UTG~\cite{utg}, and YHG~\cite{yhg} as target datasets and separately evaluated two subsets (EDSH-2 and EDSH-K) recorded in disjoint environments for the EDSH dataset.
We excluded EgoHands~\cite{egohands} from the setting of~\cite{uma} since the annotation protocol is different.
In our experiments, we resized these images to $256\times256$ pixels. 
\magenta{The details of the datasets are described in Appendix~\ref{sec:db}.}

\tbf{Implementation details.}
We randomly sampled 10 images from the target domain's training split as style images for the stylization.
We set the hyperparameters $\alpha$ of our consensus pseudo-labeling to $0.8$.
For optimization, we used the Adam optimizer~\cite{adam} with a learning rate of $10^{-5}$, $10^{-6}$, and $10^{-5}$ on EGTEA, Ego2Hands, and ObMan-Ego, respectively.
For stylization, we used a pretrained network of~\cite{photowct} trained on the COCO dataset~\cite{lin2014microsoft}. To adapt to the synthetic hands of ObMan-Ego, we fine-tuned its decoder on ObMan-Ego and the target datasets.
For the backbone segmentation networks, we used RefineNet~\cite{refinenet} following~\cite{handwild,uma}.
\magenta{In our method and the comparison methods, we trained ImageNet-pretrained RefineNet on the (style-adapted) source dataset and then started adaptation from the (style-adapted) source domain to the target domain.}
All implementations were done by PyTorch~\cite{pytorch}.

\tbf{Ablation models.}
We present variants of the proposed method: \tbf{Proposed (FgSty)}, \tbf{Proposed (CPL)}, and \tbf{Proposed (FgSty + CPL)} consisting of our stylization, our pseudo-labeling, and combinations thereof, respectively. Our full model is \tbf{Proposed (FgSty + CPL)}. 
\blue{Our full model without the consensus pseudo-labeling loss is equivalent to \tbf{Proposed (FgSty)} (see (\ref{eq:full})).}

\tbf{Baseline methods.} We compare the performance of hand segmentation with the following methods.
\begin{itemize}
    \setlength{\parskip}{0pt}
    \setlength{\itemsep}{5pt}
    \item \tbf{Source only}: RefineNet trained on the source dataset without adaptation.
    \item \tbf{PL}: A naive pseudo-labeling method that assigns pseudo-labels to unlabeled target data by thresholding the model's prediction. We set the threshold to $0.4$ for the binary semantic segmentation.
    \item \tbf{BDL~\cite{bdl}}: A bidirectional learning method with pseudo-label training for semantic segmentation, which combines an image-to-image translation model and a segmentation model. 
    We replaced the segmentation model with RefineNet for fair comparison.
    \item \tbf{UMA~\cite{uma}}: The state-of-the-art uncertainty-guided model adaptation method in which the uncertainty of pseudo-labels is estimated through Bayesian CNN. 
    \item \tbf{Target only}: RefineNet trained on the target dataset in a supervised manner. This shows an empirical upper bound of the adaptation task.
\end{itemize}

\subsection{Results}
\tableDA

\figw{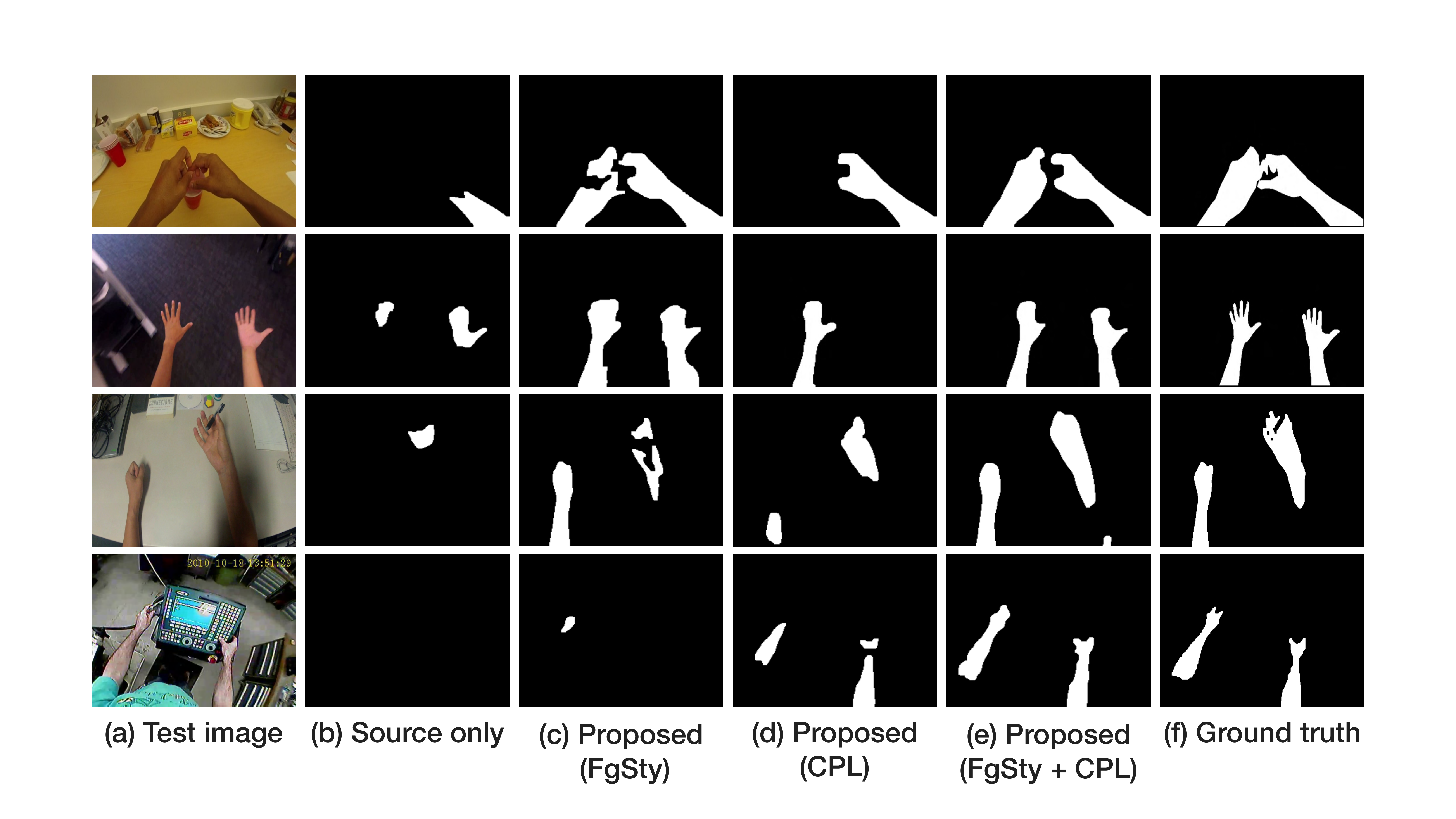}{Segmentation results. ObMan-Ego is used for the source dataset. (a) Test images and (f) ground-truth masks in GTEA, EDSH, UTG, and YHG are shown. We provide the corresponding predictions of our segmentation model from (b) source only, (c) our stylization, (d) our pseudo-labeling, and (e) our full model.}

\tbf{Real-to-real adaptation.}
Table~\ref{tbl:da} shows the results in the real-to-real setting adapted from EGTEA. \tbf{Proposed (FgSty)} performed well on all the target domains, and particularly improved by $21.00\%$ and $23.95\%$ from the source only on UTG and YHG, respectively.
This suggests our stylization effectively reduces the appearance bias in the first-person images.
\tbf{Proposed (FgSty + CPL)} outperformed PL, BDL~\cite{bdl}, and UMA~\cite{uma}.

\tbf{Sim-to-real adaptation.} 
Table~\ref{tbl:da} shows the results in the sim-to-real setting adapted from Ego2Hands or ObMan-Ego. The PL method succeeded because even target images with noisy pseudo-labels are informative in the case where the source-only prediction is not confident. Conversely, UMA failed to estimate the uncertainty of pseudo-labels and degraded their performance in the setting with the largest domain shift, \ie, in ObMan-Ego.
\tbf{Proposed (FgSty)} significantly improved the performance to $60.83\%$ and $61.05\%$, and \tbf{Proposed (FgSty + CPL)} achieved the best score of $68.74\%$ and $65.79\%$ on Ego2Hands and ObMan-Ego, respectively.

\tbf{Comparison to pseudo-labeling methods.}
In Table~\ref{tbl:da}, we compare our agreement-based pseudo-labeling (\tbf{Proposed (CPL)}) with pseudo-labeling methods based on a single model's confidence (PL, UMA). The results show PL and UMA failed when using EGTEA and ObMan-Ego, respectively. While PL naively generates the labels by thresholding the model's prediction, UMA iteratively estimates the uncertainty of pseudo-labels and weights the segmentation loss based on the estimated uncertainty for each instance. If the estimation is unreliable, such as in the ObMan-Ego setting with the largest domain shift, the model overfits highly noisy labels that should have been avoided when the estimation works as expected. The pseudo-labels generated by PL included a certain degree of noise, so the noisy supervision was harmful when the source only performed fairly well in the target domains, such as in the EGTEA setting.

In contrast, \tbf{Proposed (CPL)} performed stably in all the settings. This empirically shows that our consensus pseudo-labeling is much safer in terms of preventing the failure cases.

\tbf{Qualitative analysis.}
To confirm the adaptation ability of the proposed method, we illustrate qualitative results of the segmentation model's prediction in Fig.~\ref{fig:seg}. 
While (b) the unadapted model (source only) could not locate hands well, (c) \tbf{Proposed (FgSty)} and (d) \tbf{Proposed (CPL)} improved the hand segmentation and contributed to the improvement in different regions. For instance, the prediction of \tbf{Proposed (FgSty)} contained unsegmented parts in the hand areas of the first and third figures from the top, but the pattern seemed not to appear in the prediction of \tbf{Proposed (CPL)}. Observing the second figure from the top, \tbf{Proposed (CPL)} captured the hand shape more correctly than the model with the stylization.
(e) Our full model complementarily utilized both modules' segmentation ability and produced a sufficient quality of the segmentation masks compared to (f) ground truth.

\fig{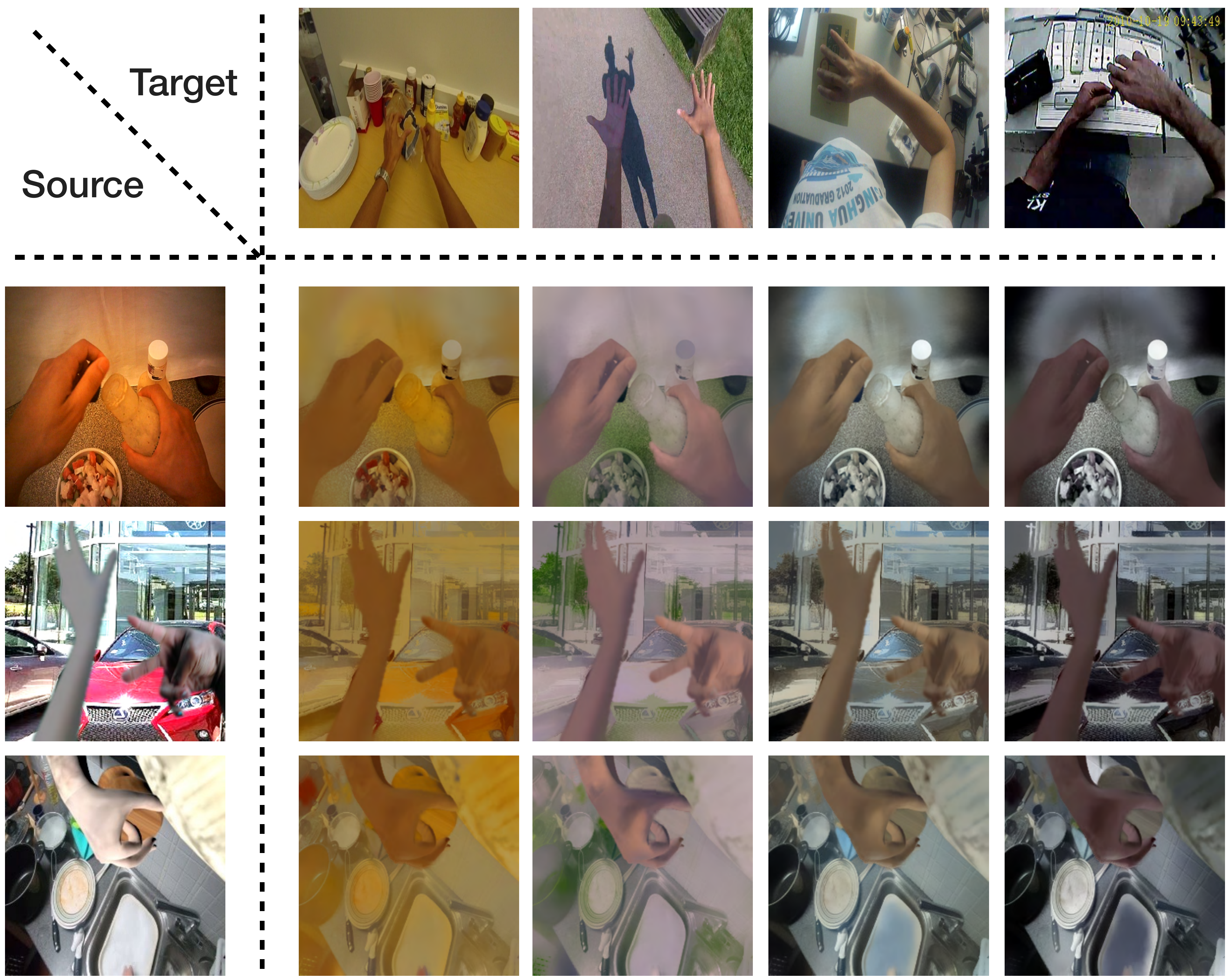}{Our stylization results. (Source: EGTEA, Ego2Hands, and ObMan-Ego, target: GTEA, EDSH, UTG, and YHG)}

\tbf{Ablation study: Foreground-aware stylization.}
Qualitative examples of our stylization are shown in Fig.~\ref{fig:vizSty}. Our stylization successfully transferred the target appearance to the style-adapted source images.

To verify the effectiveness of the separated stylization of foreground and background, we compare the segmentation performances trained on EGTEA without stylization (source only), the stylization without spatial control (unaligned), and the one with spatial control (aligned). The results are shown in Table~\ref{tbl:align}. The unaligned stylization provided little marginal gain over the source-only condition, but our stylization significantly improved across most of the target domains. However, the performance gain in the EDSH domain was unstable because its lighting condition drastically changes and the selection bias of style images strongly affected the performance.

We evaluated the impact of our foreground-aware stylization with a different number of target style images in Fig.~\ref{fig:nStyle}. 
A smaller number of target style images did not ensure sufficient variations of the target appearance and the performance was relatively weak, while the performance was almost constant when the number of style images exceeded 10. Thus, using 10 style images was a reasonable choice considering a trade-off between the annotation cost and the performance gain.

\tableUnalign
\fig{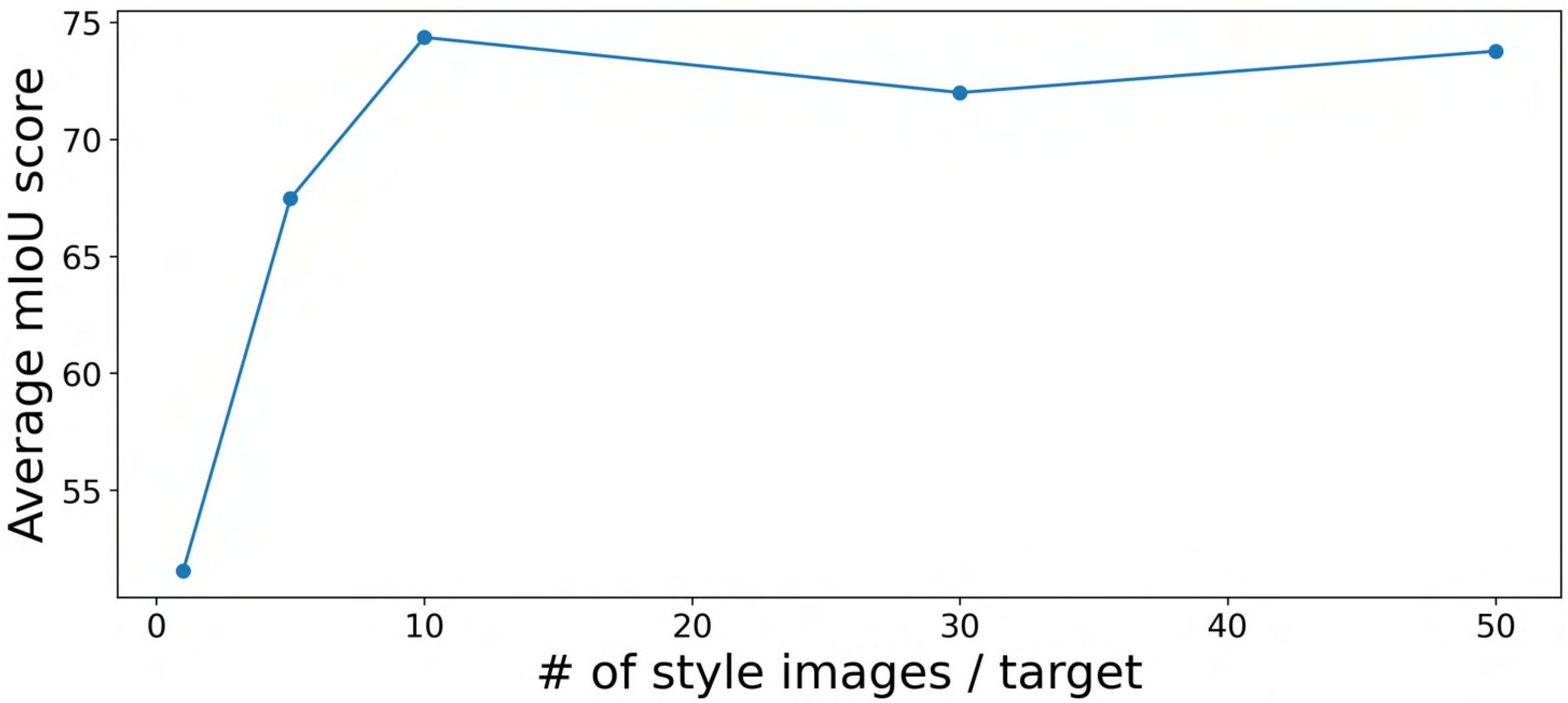}{Different number of target style images. EGTEA is used for the source dataset. We show the average performance of a segmentation model over the target datasets~\cite{gtea,edsh,utg,yhg} with a different number of style images. We set the number of the style images per target to $1$,$5$,$10$, $30$, and $50$.
}

Since reducing the appearance gap can be made by simpler baselines that have already been recognized, we compare our stylization with classical input normalizations and appearance adaptation methods with the same reference images as our stylization in Table~\ref{tbl:norm}. We added (1) the gray-scaling applied in~\cite{ego2hands}, (2) histogram equalization, (3) feature distribution matching~\cite{keepsimple}, and (4) color histogram matching~\cite{keepsimple}. These methods achieved the mean IoUs of $58.36\%$, $66.21\%$, $68.61\%$, and $66.63\%$ on average, while our stylization achieved $74.35\%$.
This demonstrates that our stylization-based approach is more useful in representing appearance and normalizing it between domains.
\tableNorm

To confirm the property of style transfer compared with recent generative domain adaptation, we tested BDL without pseudo-labeling~\cite{bdl} in the sim-to-real setting. This resulted in $31.74\%$ on Ego2Hands and $25.08\%$ on ObMan-Ego, which was worse than the source-only performance. While most pixel-level methods~\cite{cycada,learnfromsyn,dcan,bdl} use CycleGAN-based image translation~\cite{cyclegan,accr} as a backbone architecture, the training suffered significantly imbalanced data ($150$--$180$K on the source $\leftrightarrow$ $\approx500$ on targets). In contrast, style transfer exerts instance-level domain adaptation in style feature space~\cite{demysty}, so our domain translation was not affected by the data imbalance problem. This shows that style transfer is effective when the amount of target data is limited.

\fig{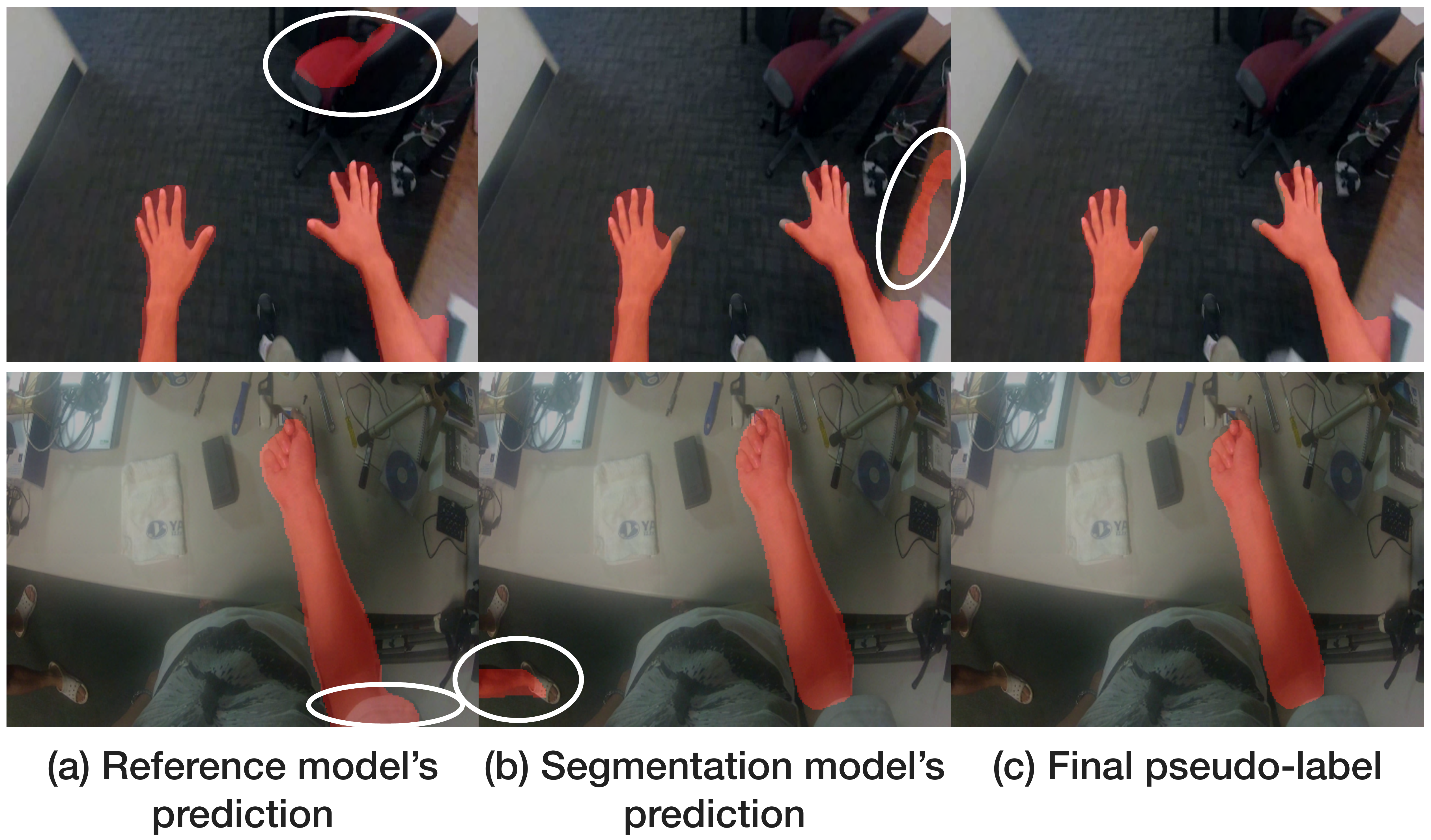}{Visualization of generated pseudo-labels. (a) Reference model and (b) segmentation model made false predictions on the chair, the table, the person's sleeve, and the bystander's leg. (c) Final pseudo-labels suppressed the mislabels by taking an intersection between the two outputs.}
\fig{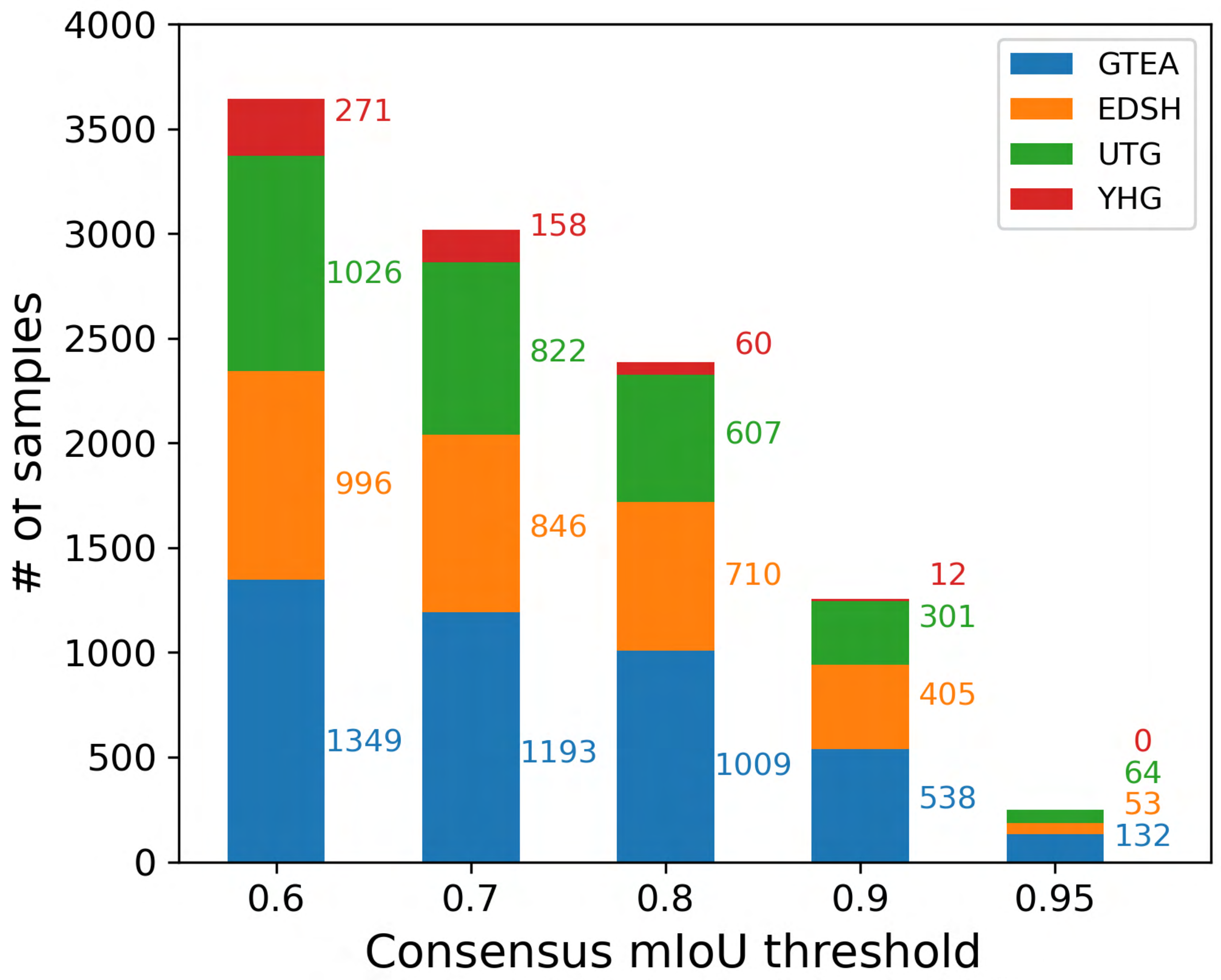}{Number of consensus pseudo-labels. We show the number of the agreed pseudo-labels by our pseudo-labeling.}
\fig{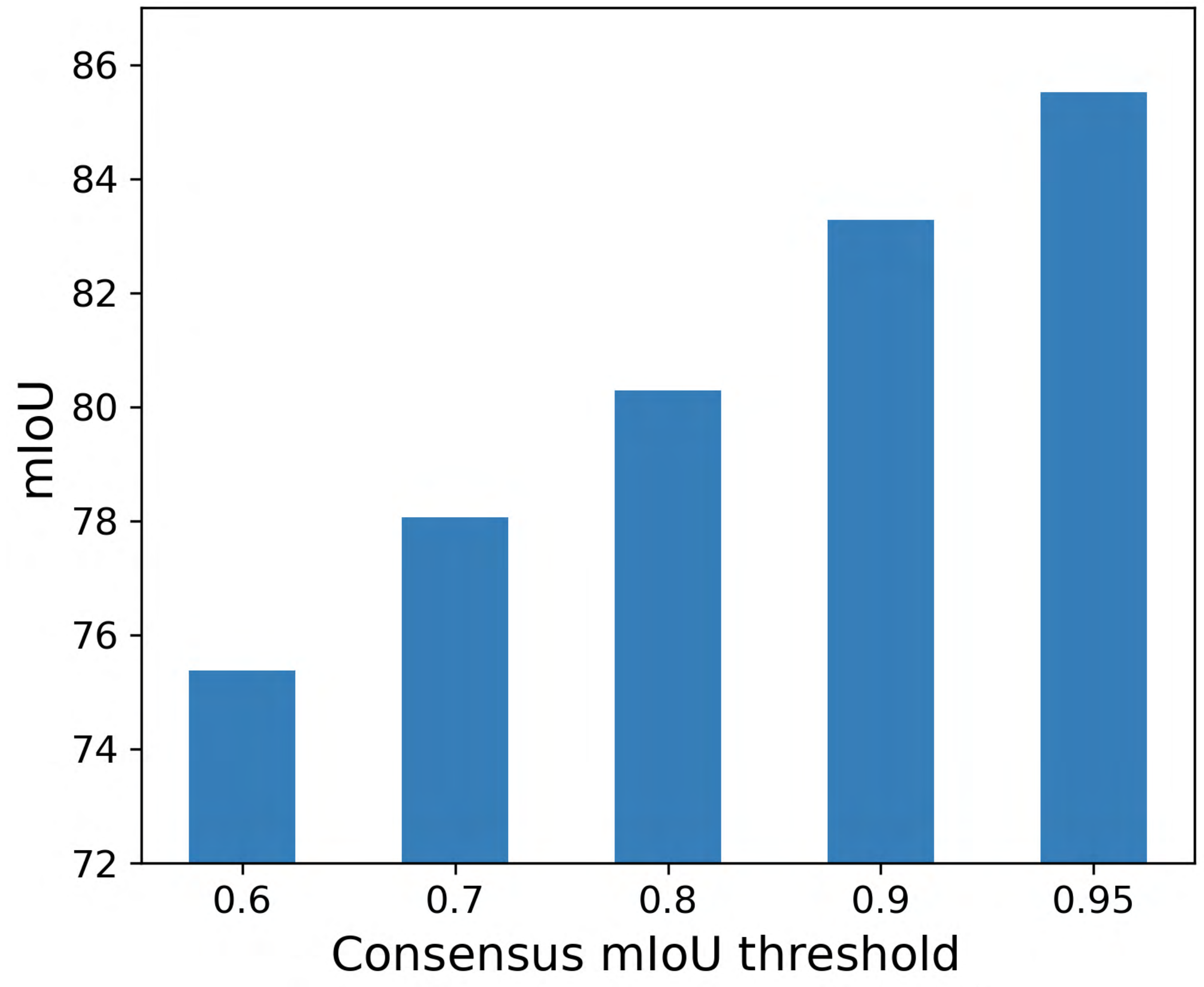}{Quality of consensus pseudo-labels. We show mean IoU measured by the agreed pseudo-labels and their ground truth.}

\tbf{Ablation study: Consensus pseudo-labeling.}
The qualitative results of consensus pseudo-labeling are shown in Fig.~\ref{fig:cplPred}.
While the baselines incorrectly gave labels on backgrounds such as the chair (top left), the table (top middle), and the bystander's leg (bottom middle), our consensus mechanism suppressed such false predictions by utilizing the reference model’s segmentation ability.

To investigate the effect of the hyperparameter $\alpha$ of our consensus pseudo-labeling, we present the number of pseudo-labels available in an epoch and their quality compared with ground truth in Figs.~\ref{fig:cplDist} and~\ref{fig:cplQual}, respectively. As the threshold increases, the number of the agreed pseudo-labels decreased, but the quality steadily increased.
\blue{
We found that the number of the pseudo-labels in YHG was very limited when the value $\alpha$ was equal to $0.9$ or above. To balance the number and quality of the pseudo-labels, we set the value $\alpha$ to $0.8$ in all experiments.
}

\tableSpeed
\red{
Since we utilized an additional network in the adaptation training of our pseudo-labeling, we compare the training speed with other pseudo-labeling methods in Table~\ref{tbl:speed}. Here, our pseudo-labeling method used two GPUs for running the two networks in parallel. While the computational time of naive training on a dataset and the naive pseudo-labeling method (PL) were 394 and 653, respectively, our method (\tbf{Proposed (CPL)}) with two GPUs ran a training update faster than PL, though the total computational cost was doubled. Due to estimating the uncertainty map of an instance with several forward calculations~\cite{uma}, UMA took the longest time of 841.
}

\tbf{Ablation study: Size of the source data.}
\tableSSize
\red{
To reveal how the size of the labeled source data affects the adaptation performance, we experimented with various sizes of the labeled data in Table~\ref{tbl:ssize}. While the source-only performance gradually degraded when changing the size of EGTEA from 13K to 500, \tbf{Proposed (FgSty + CPL)} demonstrated robustness to the change in the source data size and achieved high performance even with a small amount of the labeled source data.
}

\subsection{Additional Experiments}
\tbf{Integration with adversarial adaptation.}
Since only a few target images are used to incorporate their styles into the style-adapted dataset and update the model with their pseudo-labels, \tbf{Proposed (FgSty + CPL)} cannot take advantage of all unlabeled target samples. To align feature distributions between the domains by utilizing unlabeled target samples, we further combine adversarial adaptation with both models $M$ and $R$. 
We adapted a typical domain discriminator~\cite{dann,adda} to discriminate inputs pixel wisely. The pixel-wise discriminator classifies whether extracted features come from the (style-adapted) source or target dataset. The adversarial training is facilitated by a gradient reversal layer~\cite{dann}.

Table~\ref{tbl:adv} shows the results of our method with adversarial adaptation.
The adversarial training was effective in the sim-to-real setting with a large domain shift, \eg, when using ObMan-Ego as the source dataset. During adaptation in \tbf{Proposed (FgSty + CPL)}, the performance of the reference model (source only) was weak and fixed, but \tbf{Proposed (FgSty + CPL + Adv)} updated the reference model to adapt to the target domain. The better reference model takes larger areas of the agreement with the segmentation model, which boosts the adaptation performance.

When taking consensus from the networks trained on the same source dataset with adversarial adaptation (\tbf{Proposed (CPL + Adv)}), the local feature alignment by the discriminator promoted to learn similar representation and the two networks tended to agree upon false prediction. This had an adverse impact on giving pseudo-labels based on the agreement. Nevertheless, when using stylized and unstylized source datasets in \tbf{Proposed (FgSty + CPL + Adv)} and having the two networks predict data from different viewpoints, such conflict between our pseudo-labeling and adversarial adaptation was avoided and achieved further improved results.

\tableAdv

\tbf{Extension to multi-target domain adaptation.}
Although single-target domain adaptation methods are sufficient when a well-defined target domain exists, target domains often consist of diverse distributions in the wild~\cite{blendmtda}.
We therefore further validate our method in a multi-target domain adaptation setting~\cite{mtdainfo,blendmtda} that has not been explored much. We aim to simultaneously adapt to multiple target domains by one-time adaptation training.
Unlike GAN-based adaptation methods~\cite{cycada,learnfromsyn,dcan,bdl}, the style transfer enables the translation of an image among diverse scenes~\cite{deepphoto,photowct}, so we can easily extend to the multi-target setting.
In the stylization, 10 target style images were collected per target and the target images used in training were uniformly sampled from each target dataset.

\tableMTDA
The performance in the multi-target domain adaptation setting is shown in Table~\ref{tbl:mtda}.
Although a performance degradation could be expected when pseudo-labels in a domain are incorrectly estimated and negatively affected adaptation performance in the other domains, our method in the multi-target setting achieved equal or sometimes better performance to that in the single-target setting. 
This is because other domain knowledge is informative if the pseudo-labels correctly represent the target labels. With the combination of multi-target adaptation and adversarial adaptation explained previously, \tbf{Proposed (FgSty + CPL + Adv)} demonstrated a further performance gain.

\tbf{Extension to multi-auxiliary domain generalization.}
When the test data is not available during training due to some restriction, \eg, pertaining to resources or privacy issues, generalizing to the test domain is important. We experimented with our method in a multi-auxiliary domain generalization setting~\cite{stydg}.
Here, we chose a domain from the target domains as a test domain.
We call the other target domains auxiliary domains and utilized them as the domains to adapt during training.

\tableDG
The performance in the domain generalization setting is shown in Table~\ref{tbl:dg}.
Without access to the test domain, our method achieved a significant improvement of $59.85\%$ on average.
Adaptation to the reference domains generalized to unseen domains when environmental conditions were partially shared among domains.
However, the performance on YHG was poor since half of the YHG videos were collected in a machine shop that does not appear in the other datasets.
\section{Discussion}
\tbf{Domain divergence relaxation.}
Our foreground-aware stylization can be thought of as a solution to relax the domain divergence.
Theoretically, \cite{datheory} showed that the expected error in the target domain is bounded by its source domain error and a divergence measure between the source and target domain. If model capacity is sufficient, the error in the source domain is expected to be small. Hence, the performance of domain adaptation heavily depends on the divergence. As shown in Table~\ref{tbl:da}, the comparison methods' performances based on their own model's confidence (PL, UMA~\cite{uma}, and BDL~\cite{bdl}) are sensitive to the problem setting and handling a large domain shift, \eg, in Ego2Hands and ObMan-Ego, is difficult. In this case, we need to introduce specific heuristics, such as the mask separation in our stylization.
With a few target labels, our stylization performed powerfully in reducing the domain divergence.

\tbf{Leveraging multi-domain knowledge.}
We demonstrated multi-target domain adaptation and multi-auxiliary domain generalization that few works can address. The principle is to generalize within the multiple target (reference) domains and predict data in the target (unseen) domain using the attained generalization ability. Our experiments show that the multi-target adaptation was equally as effective as the single-target adaptation and the generalization scheme had a certain effect under the condition that the reference and test domains are similar. Considering that the training cost of single-target adaptation is proportional to the number of the target domains, adaptation and generalization using multiple domain knowledge are more practical.

\tbf{Adapting to hands occluded by objects.}
\red{
As in~\cite{handwild}, our work still cannot address segmenting a person's hand holding an object well. Our future work will involve adapting to the person interacting with objects in the environment, such as the scenes of opening a shelf, holding a cup, and occluding their hands by clothing or a wristwatch.
}

\section{Conclusion}
In this work, we tackled the problem of domain adaptation in first-person hand segmentation.
We proposed two techniques: foreground-aware stylization and consensus pseudo-labeling.
The foreground-aware stylization is simple but effective in reducing the appearance bias of first-person images. 
The consensus pseudo-labeling discreetly selects reliable pseudo-labels even when the domain shift is large. Our method delivered state-of-the-art results for real-to-real and sim-to-real adaptation. 
The combination of our stylization and pseudo-labeling not only performed stably in settings with various source-target pairs, but can also be applied to adapting to multiple target domains and generalizing to unseen domains when similar auxiliary domains are available.


\appendix
{
\section*{Appendix}
\subsection{Dataset Properties}\label{sec:db}
\magenta{
\tbf{EGTEA~\cite{egtea}.} 
The Extended GeorgiaTech Egocentric Activity (EGTEA) dataset contains $29$ hours of egocentric videos. These videos record meal preparation performed by $32$ subjects in a naturalistic kitchen environment. $13,847$ images are labeled with hand masks. 
In our experiment, we used all the images and labels as the source dataset in domain adaptation.}

\magenta{
\tbf{Ego2Hands~\cite{ego2hands}.} 
The Ego2Hands dataset is a large-scale synthetic hand dataset rendered from a massive amount of annotated right hands and their horizontally flipped counterparts. The dataset consists of a training set with $188,362$ images and a test set with $2,000$ images. The foreground hands are collected from $22$ participants with diverse skin colors and hand features, and the backgrounds are collected from the DAVIS datasets~\cite{davis2016,davis2017}.
}

\magenta{
\tbf{ObMan-Ego.} 
The ObMan-Ego is a large-scale synthetic hand dataset with egocentric scenes in which the simulated hands are provided by ObMan~\cite{obman}. Training, validation, and testing sets contain $150,000$, $6,500$, and $6,500$ images, respectively. The ObMan is generated by Graspit~\cite{graspit}, an automatic robotic grasping software and ShapeNet~\cite{shapenet} object models. We rendered ObMan with the backgrounds of two large-scale egocentric videos, EPIC-KITCHENS100~\cite{epic100} and Something-Something~\cite{something}. To collect egocentric scenes without hands, we eliminated frames in which hands appeared by using a hand detector~\cite{shan20hand} and then used the remained frames for the rendering.
}

\magenta{
\tbf{GTEA~\cite{gtea}.} 
The GeorgiaTech Egocentric Activity (GTEA) consists of $28$ egocentric videos recording seven daily activities performed by four subjects. $663$ images are annotated with hand masks. We followed the data split as in~\cite{handwild,uma}, where images of the subjects $1$, $3$, and $4$ were used as a training set and the rest as a testing set.
}

\magenta{
\tbf{EDSH~\cite{edsh}.} 
The CMU EDSH dataset contains three egocentric videos, including EDSH1, EDSH2, and EDSH-Kitchen, which are recorded in indoor and outdoor environments. We adopted the same data split as in~\cite{edsh,uma}. $442$ labeled images in EDSH1 were used as a training set. $104$ labeled images in EDSH2 and $197$ labeled images in EDSH-Kitchen were used as two separate testing sets.
}

\magenta{
\tbf{UTG~\cite{utg}.} 
The University of Tokyo Grasping (UTG) dataset consists of $50$ egocentric videos and captures $17$ different types of hand grasps performed by five subjects. We used the annotation of hand masks on $857$ images provided by~\cite{uma} and randomly split them into training and testing sets with a ratio of $75\%$ and $25\%$, respectively. 
}

\magenta{
\tbf{YHG~\cite{yhg}.}
The Yale Human Grasping (YHG) dataset provides daily observations of human grasping behavior in unstructured environments. It consists of $27.7$ hours of egocentric videos recorded by two machinists and two housekeepers during their daily work. Following~\cite{uma}, we used the annotation of hand masks on $488$ images and randomly split them into training and testing sets with a ratio of $75\%$ and $25\%$, respectively.
}
}

\EOD

\end{document}